\documentclass[conference]{IEEEtran}
\IEEEoverridecommandlockouts

\usepackage{amsmath,amssymb,bm}
\usepackage[hidelinks]{hyperref}
\usepackage{graphicx}

\begin{document}
\title{\vspace{4mm} Spinning Quadrotor: Hover Thrust Augmentation with Passive Lifting Surfaces}

\author{
    Aniketh Parkala \and Harikumar Kandath%
    \thanks{Both authors are with the Robotics Research Center (RRC), 
    IIIT Hyderabad, India (e-mail: anikethparkala@gmail.com; harikumar.k@iiit.ac.in). 
   }
}
\maketitle

\begin{abstract}
Conventional multirotor aerial vehicles actively suppress yaw rotation during hover, expending power to maintain a fixed heading despite the fact that yaw regulation is not required for force balance or altitude control. This paper challenges that paradigm by proposing a spinning quadrotor architecture that intentionally operates at a sustained yaw rate, converting power traditionally spent on yaw regulation into useful aerodynamic effects. A dynamic model of the spinning quadrotor is developed, analysis for low Re range is conducted to choose an airfoil for lifting surfaces. Preliminary hardware tests show a 22\% reduction in thrust required. These findings suggest that intentional yaw rotation, rather than being suppressed, can be exploited as a design mechanism for efficient and robust multirotor flight.
\end{abstract}

\section{Introduction}
\subsection{Motivation}
Multirotor vehicles typically suppress yaw during hover using differential rotor torques. This work instead considers sustained intentional yaw rotation as an alternative operating point. By canting all rotors in a common direction, the vehicle converges to a steady spin while regulating altitude and reduced attitude. The resulting tangential airflow enables lift generation from radially mounted aerodynamic surfaces, reducing propeller thrust demand.

Various use cases of spinning drones have been given below.
\subsubsection{ Fault-tolerant control (FTC) for quadrotors via spinning}
\noindent A substantial body of work studies spinning as a \emph{recovery mechanism} for conventional quadrotors after actuator loss. In these ``relaxed hover'' approaches, yaw regulation is intentionally sacrificed so that the remaining actuators can regulate thrust magnitude and direction. Mueller and D'Andrea\cite{mueller_stability_2014} derive periodic spinning solutions that retain position control even with the complete loss of one, two, or three propellers . Freddi et al. and Lanzon et al.\cite{freddi_feedback_2011,lanzon_flight_2014} use feedback linearization (and robust variants) to stabilize the vehicle under single-rotor failure by converging to a constant-spin equilibrium . More recently, Ke et al.\cite{ke2023uniform} propose a uniform passive FTC that treats rotor faults as a lumped disturbance, avoiding controller switching and explicit fault detection . Practical deployment also raises sensing challenges during high-spin flight; Sun et al.\cite{sun_autonomous_2021} study visual-inertial estimation under rapid spinning and highlight the limitations of standard cameras due to motion blur .

\subsubsection{ Novel monocopter and samara-inspired designs}
\noindent A second theme departs from the quadrotor form factor and instead designs vehicles whose \emph{primary} lift and stability mechanisms are inherently rotation-driven. Examples include foldable single-actuator monocopters inspired by samara seeds \cite{win_design_2021}, bioinspired revolving-wing drones that exploit unsteady aerodynamics for high power loading \cite{bai2022bioinspired}, and modular monocopter systems that can connect and separate in flight \cite{cai_cooperative_2022}. Related analysis on mono-spinners shows that asymmetric layouts with deliberate tilt can reduce power consumption while maintaining stable hover \cite{hedayatpour2017optimal}.
\begin{figure}
    \centering
    \includegraphics[width=1\linewidth]{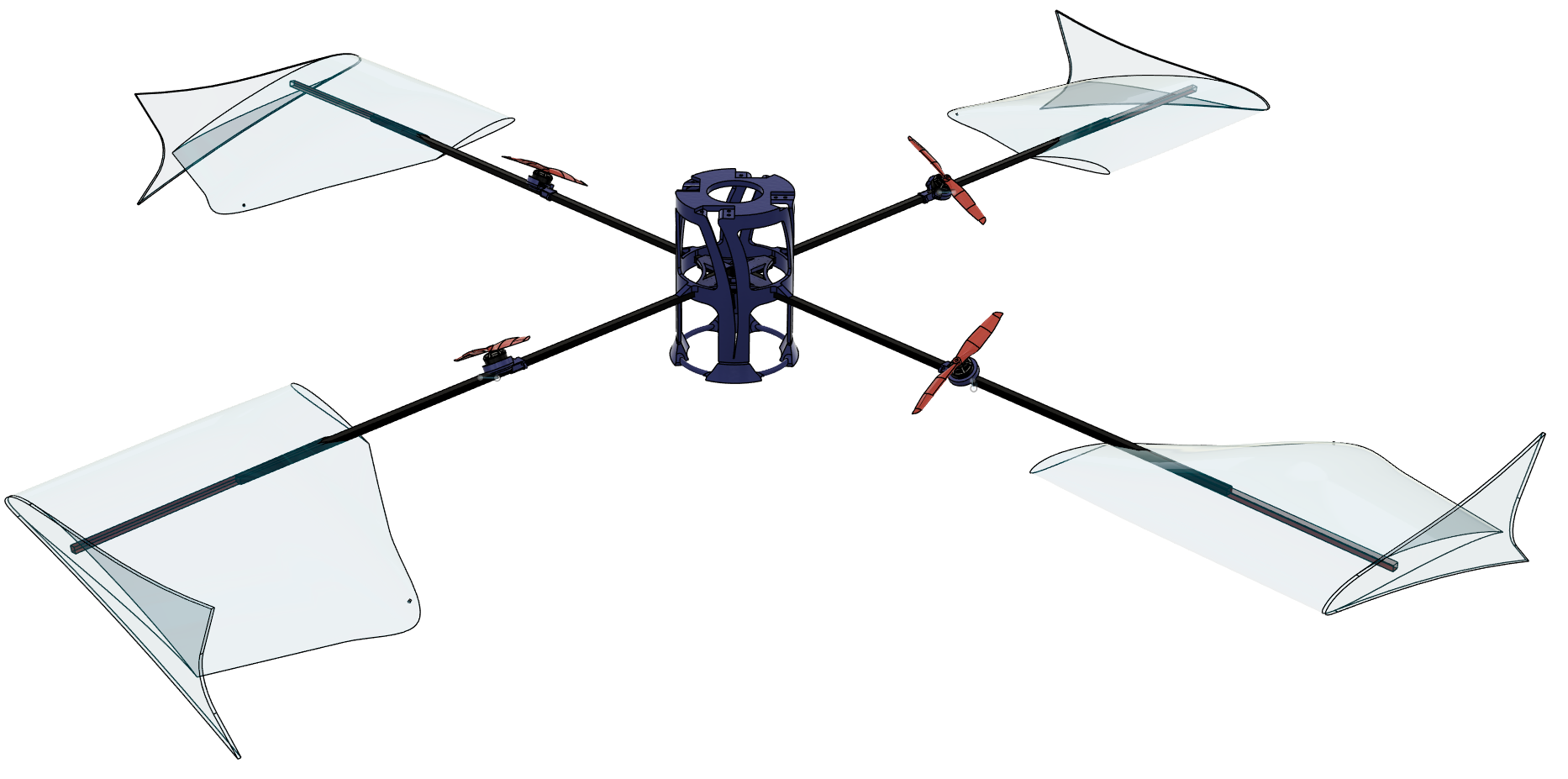}
    \caption{CAD model of configuration 2. All motors are canted 10 degrees}
    \label{fig:config2cad}
\end{figure}

\subsubsection{ Hybrid and transformable rotary platforms}
\noindent A third line of research explores hybrid or mechanically transformable rotary platforms that bridge hovering and efficient forward flight. Low et al. \cite{low_design_2017} present a transformable hovering rotorcraft that transitions between whole-body spinning hover and fixed-wing cruise . Orsag et al. \cite{orsag_spincopter_2013} analyze spincopters in which thrust pulsation produces horizontal motion while the craft spins. In contrast to under-actuated spinning strategies, over-actuated tilt-arm quadrotors can decouple attitude and translation via thrust vectoring to achieve full pose tracking \cite{invernizzi2018full}.

Building on these three themes, this paper focuses on spinning as an \emph{intentional} hover operating point for a quadrotor-scale platform, primarily motivated by efficiency: replacing continuous yaw suppression with steady rotation, exploiting spin-averaging of disturbances, and enabling lift augmentation from aerodynamic surfaces during hover.

The contributions of this work \footnote{Code:\url{https://github.com/pyromania99/spinning-quadrotor-hover}} are as given below.
\begin{enumerate}
  \item We propose a spinning-quadrotor hover architecture that replaces yaw suppression with intentional steady rotation and enables aerodynamic lift augmentation with passive lifting surfaces.
  \item Design of an optimal airfoil for the lifting surfaces in the low Reynolds number (Re) regime, based on glider airfoil and XFLR analysis.
  \item Multi-fidelity validation via simulation (BEMT), CFD, and hardware experiments which show a promising increase of 22\% of lift when compared to conventional quadrotors.
\end{enumerate}

\section{Problem Premise and Operating Conditions}
Conventional quadrotor hover is defined as the equilibrium condition given by Eq.~\eqref{eq:quadrotor_hover}.
\begin{equation}
\dot\omega_z = 0, \quad \bm{\omega} = 0
\label{eq:quadrotor_hover}
\end{equation}
Where $\omega_z$ (rad/s) is the yaw rate, $\bm{\omega}$ (rad/s) is angular velocity, with
the thrust balancing the weight, and all angular rates are regulated to zero. In contrast, this work considers a modified hover equilibrium
\begin{equation}
\dot\omega_z = 0, \quad  \omega_z \neq 0,
\end{equation}
in which altitude and reduced attitude (roll and pitch) are regulated while yaw converges to a nonzero steady rate. Yaw is therefore treated as an operating parameter rather than a disturbance to be rejected.
At steady yaw rate $\omega$, a point located at radius $r$ experiences tangential velocity
\begin{equation}
V(r) = \omega_z r.
\end{equation}
where $r$ (m) is radial distance from the vehicle center.

Thus, radially mounted lifting surfaces experience effective airflow even in zero freestream conditions. The generated lift scales as shown in Eq.~\eqref{eq:lift}
\begin{equation}
L = \tfrac{1}{2}\rho (\omega^2 r^2) S C_L,
\label{eq:lift}
\end{equation}

where $\rho\ (kg/m^3)$ is the air density, $\omega$ (rad/s) is the steady yaw rate, 
$r$ (m) is the radial position of the lifting surface, $S$ (m$^2$) is the wing 
planform area, and $C_L$ is the lift coefficient.
This suggests that intentional spinning may augment lift and reduce propeller thrust demand. However, this regime is bounded by control, aerodynamic, structural, and safety constraints.
\subsection{Constraints}
\paragraph{Yaw-rate authority and control bandwidth.}
Motor bandwidth limits the authority at higher yaw rates. As steady yaw increases, rate-loop saturation, unaccounted cross-axis gyroscopic coupling ($\bm{\omega}\times I\bm{\omega}$), reduced mixer authority, and estimator bandwidth limitations degrade reduced-attitude performance. These effects are primarily control- and bandwidth-limited rather than arising from structural imbalance or rigid-body natural frequency constraints. Experimentally, stable spinning hover is maintained up to 
\begin{equation}
\omega \approx 8\text{--}10~\mathrm{rad/s}.
\end{equation}
Beyond this range, oscillatory behavior and loss of roll--pitch robustness are observed. While nonlinear or spin-aware controllers (as used in fault-tolerant spinning literature) could permit higher rates, the present implementation is intentionally limited to the standard PX4 cascaded framework. Accordingly, $\omega \leq 10~\mathrm{rad/s}$ defines the practical upper. The standard PX4 controller is used without structural modification; to accommodate continuous yaw rotation, the yaw reference is offset so that roll and pitch commands remain defined in a fixed inertial (world) frame.
The total mass is constrained to $m < 1~\mathrm{kg}$ for safe experimental validation.
\paragraph{Aerodynamic velocity and Reynolds regime.}
meaningful aerodynamic lift must be generated within the allowable structural radius. The radial arm length was therefore extended to $l \leq 0.65~\mathrm{m}$, where $l$ (m) is the motor arm length, beyond which structural flexibility and vibration modes increase noticeably during spin. At this maximum radius, the achievable tangential velocity is given by Eq.~\eqref{eq:V_max}.
\begin{equation}
V_{\max} = \omega l \approx 6.5~\mathrm{m/s}.
\label{eq:V_max}
\end{equation}

To obtain Reynolds numbers sufficiently high for practical lift generation within this velocity bound, relatively large chord lengths are required. For chord values in the range $c \approx 0.20$--$0.30~\mathrm{m}$, the resulting Reynolds number lies in the low-Re regime,
\begin{equation}
\mathrm{Re} = \frac{\rho V c}{\mu} \sim 10^5.
\end{equation}
where $c$ (m) is the chord length,  and $\mu$ (Pa$\cdot$s) is the dynamic viscosity of air. Accordingly, airfoil selection is constrained to profiles optimized for low-Re operation, making glider-inspired airfoils a natural and physically motivated choice. Together, these constraints define a bounded but experimentally realizable spinning-hover regime within which aerodynamic lift augmentation can be evaluated.
\subsection{Rotor  Forces}
Each rotor produces thrust \begin{equation}
f_i = k_f \omega_i^2,
\label{eq:rotor_thrust}
\end{equation}
where $f_i$ (N) is rotor thrust, $k_f$ (N$\cdot$s$^2$) is the thrust coefficient,
and $\omega_i$ (rad/s) is rotor angular speed.
The resulting force vector is defined in Eq.~\eqref{eq:resulting force}.
\begin{equation}
\mathbf F_i = f_i\big(\sin\alpha\,\hat t_i + \cos\alpha\,\hat z\big),
\label{eq:resulting force}
\end{equation}
where $\hat t_i$ is the tangential unit vector in the $xy$-plane , and $\alpha$ (rad) is the motor cant angle.

\subsection{Total Forces and Moments}
The net vertical force and control moments equations are given in Eq. (\ref{forces_eqn_begin})-(\ref{forces_eqn_end})
\begin{align}
F_z &= \sum_{i=1}^4 f_i \cos\alpha, \label{forces_eqn_begin}\\
M_\phi &= l \cos\alpha (f_1 + f_2 - f_3 - f_4), \\
M_\theta &= -l \cos\alpha (f_1 - f_2 + f_3 - f_4), \\
M_\psi &= \sum_{i=1}^4 \big(k_\tau \omega_i^2 + f_i \sin\alpha \, l \big).
\label{forces_eqn_end}
\end{align}

where $M_\phi$, $M_\theta$, and $M_\psi$ (N$\cdot$m) denote the roll, pitch, and yaw moments about the body-fixed $x$, $y$, and $z$ axes, respectively; and $k_\tau$ (N$\cdot$m$\cdot$s$^2$) is the propeller torque coefficient.

\subsection{Dynamics of motion}

The dynamics of motion are given in Eq. (\ref{dynamics_eqn_begin})-(\ref{dynamics_eqn_end})
\begin{align}
m(\dot{\mathbf v} + \bm{\omega}\times\mathbf v) &= mR^\top \mathbf g + \sum_i \mathbf F_i, \label{dynamics_eqn_begin}\\
I\dot{\bm{\omega}} + \bm{\omega}\times(I\bm{\omega}) &= \mathbf {M}
\label{dynamics_eqn_end}
\end{align}
with angular rates $\bm{\omega}=[\omega_x,\omega_y,\omega_z]^\top$.

Where $\mathbf v$ (m/s) is body-frame velocity, $I$(kg$\cdot$m$^2$) is the inertia matrix, where $R \in SO(3)$ is the body-to-inertial rotation matrix, $\mathbf g=[0,0,-g]^\top$ and $\mathbf M$ (N$\cdot$m) is the net moment vector.

\section{Hardware Configuration and Design}
To maintain a non zero-yaw rate against wing drag, all of the motors are mounted spinning and canted in the same direction, to add to the torque produced by rotating the propellers in the same direction. The angle at which they are canted is fixed precisely by 3d printing the motor mounts and affixing them to square carbon arms.
The arm length of the design has been extended to 0.5 m to support the lift surfaces. The lift surface chosen is based on a custom airfoil which we will explore in section(\ref{subsec:custom_airfoil}). Winglets have been added to reduce any wake and edge vortex effects. Two distinct airframe configurations were evaluated to examine the relationship between motor placement, thrust vector orientation, and resulting force--moment generation.
\paragraph{Configuration I: Lever-Arm-Optimized Asymmetric Layout}
The first configuration employs an asymmetric quad-motor arrangement designed to enhance rotational authority. Two motors are positioned along the secondary body axis at a radial distance of 0.5~m from the vehicle center of mass and are tilted outward by 10$^\circ$. The increased lever arm is intentionally selected to amplify the generated torque for a given thrust level, thereby optimizing the configuration for spinning motion.

\begin{figure}
    \centering
    \includegraphics[width=1\linewidth]{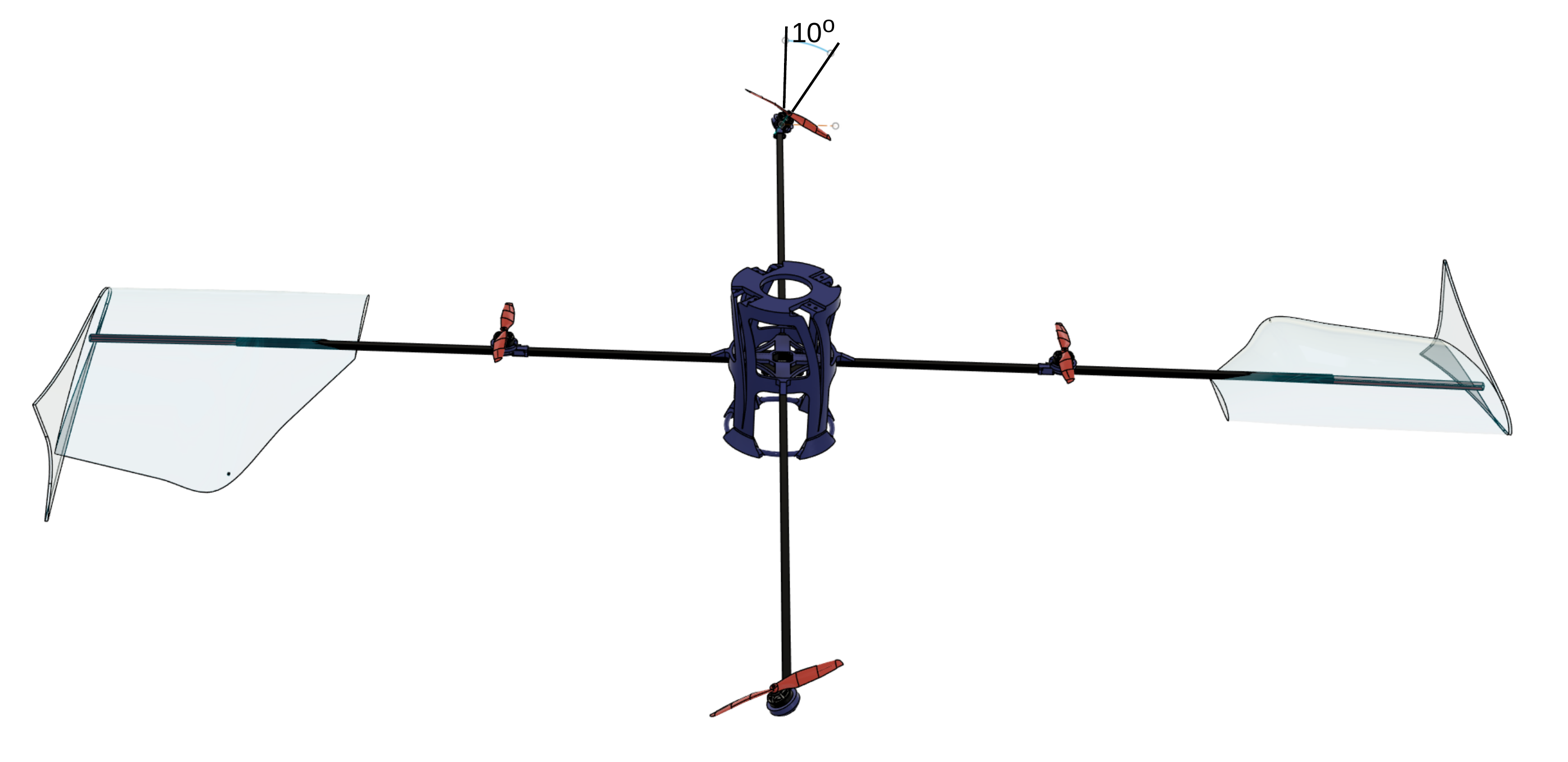}
    \caption{CAD model of configuration 1, with only 2 motors canted on non-wing arms (exaggerated motor tilt for visualization)}
    \label{fig:config1cad}
\end{figure}

Along the primary axis, the remaining two motors are located at a radial distance of 0.3~m, with two wings further outboard, spanning radially from 0.4~m to 0.65~m. This configuration is intended to approximate the upper bound of isolated wing thrust generation by reducing aerodynamic interference effects between lifting surfaces. A model of Configuration~I is shown in Fig.~\ref{fig:config1cad}.
\paragraph{Configuration II: Symmetric Fully Tilted Distributed Wing Layout}
The second configuration adopts a more symmetric geometry to evaluate the trade-off between increased lifting surface area and the associated aerodynamic interference effects. All four motors are mounted at a uniform radial distance of 0.3~m and are tilted by 10$^\circ$. One wing is aligned along each motor arm, extending radially from 0.4~m to 0.65~m.

This configuration removes lever-arm-induced bias and isolates the effects of uniform thrust vectoring and distributed lift generation. A schematic of Configuration~II is shown in Fig.~\ref{fig:config2cad}.

\section{Airfoil Selection}
\label{sec:airfoil_selection}
Since the controllable steady yaw rate is limited to approximately
$\omega \approx 8$--$12~\mathrm{rad/s}$ and the wing radius is
$r \leq 0.65~\mathrm{m}$, the characteristic tangential flow velocity
$V(r) \approx \omega r$, combined with chord lengths on the order of
$c \approx 0.25$--$0.30~\mathrm{m}$, places the lifting surfaces firmly
within the low-Reynolds-number regime. Accordingly, candidate airfoils
were selected from glider and low-Reynolds-number airfoil families and
evaluated over the relevant $(\alpha, \mathrm{Re})$ operating space.

\subsection{Candidate Airfoils and Screening Procedure}
\label{subsec:airfoil_candidates}
The candidate airfoils considered were AG-14, RG-14, RG-15, S1223, SD7032, SD7037, SD7062, and NACA 4412. 
For each airfoil, polar data were generated across multiple Reynolds numbers and angles of attack. Aerodynamic efficiency, quantified by \(C_l/C_d\)
 within the expected operating envelope, served as the primary screening metric to reduce the candidate set prior to detailed operating-point optimization.

\subsection{Custom Airfoil Generation}
\label{subsec:custom_airfoil}
\begin{figure}
    \centering
    \includegraphics[width=1\linewidth]{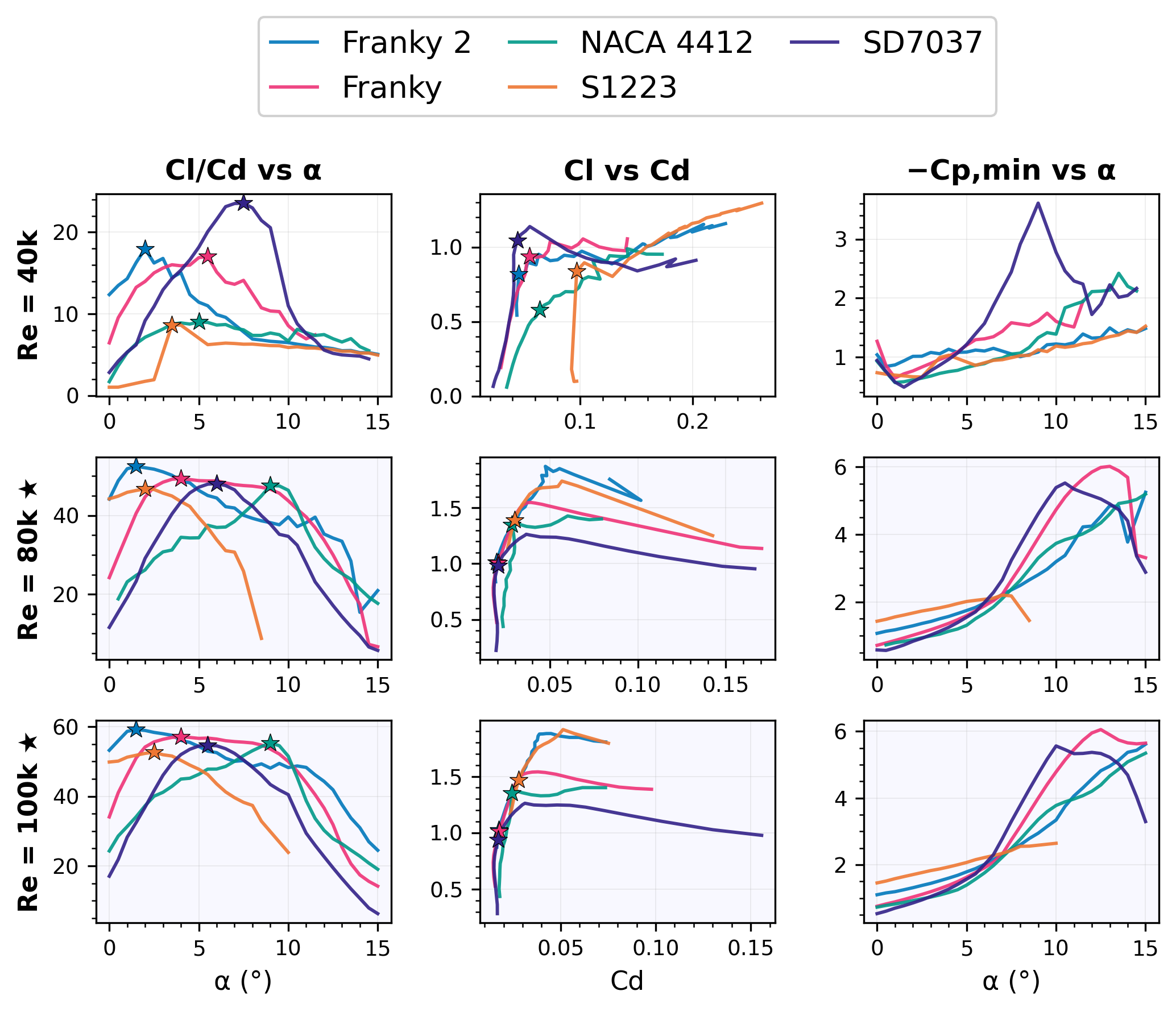}
    \caption{Airfoil characteristics comparison}
    \label{fig:airfoil_comparison}
\end{figure}

A custom airfoil, denoted \emph{Franky}, was generated by interpolating the geometries of S1223 and SD7037 with a 26\% blend ratio (S1223--SD7037). This interpolation was chosen to combine the high-lift characteristics of S1223 with the high \(C_l/C_d\) at low-$\mathrm{Re}$ of SD7037. 
A second candidate, Franky-2 (60\% blend), demonstrated competitive performance but exhibited optimal behavior over a narrower range of angles of attack and was therefore not selected $\alpha$ values Fig~\ref{fig:airfoil_comparison}

\subsection{Reynolds number-Specific Operating-Point Optimization}
\label{subsec:airfoil_selection}

Airfoil performance was evaluated over the Reynolds range Re = 0.040–0.090 $\times 10^6$, corresponding to the operational envelope of the lifting surfaces. As no aerodynamic surfaces exist in the hub region, all blade sections operate entirely within this range during steady spinning-hover. 

\begin{figure}[t]
    \hspace*{-3mm}
    \centering
    \includegraphics[width=1.00\linewidth]{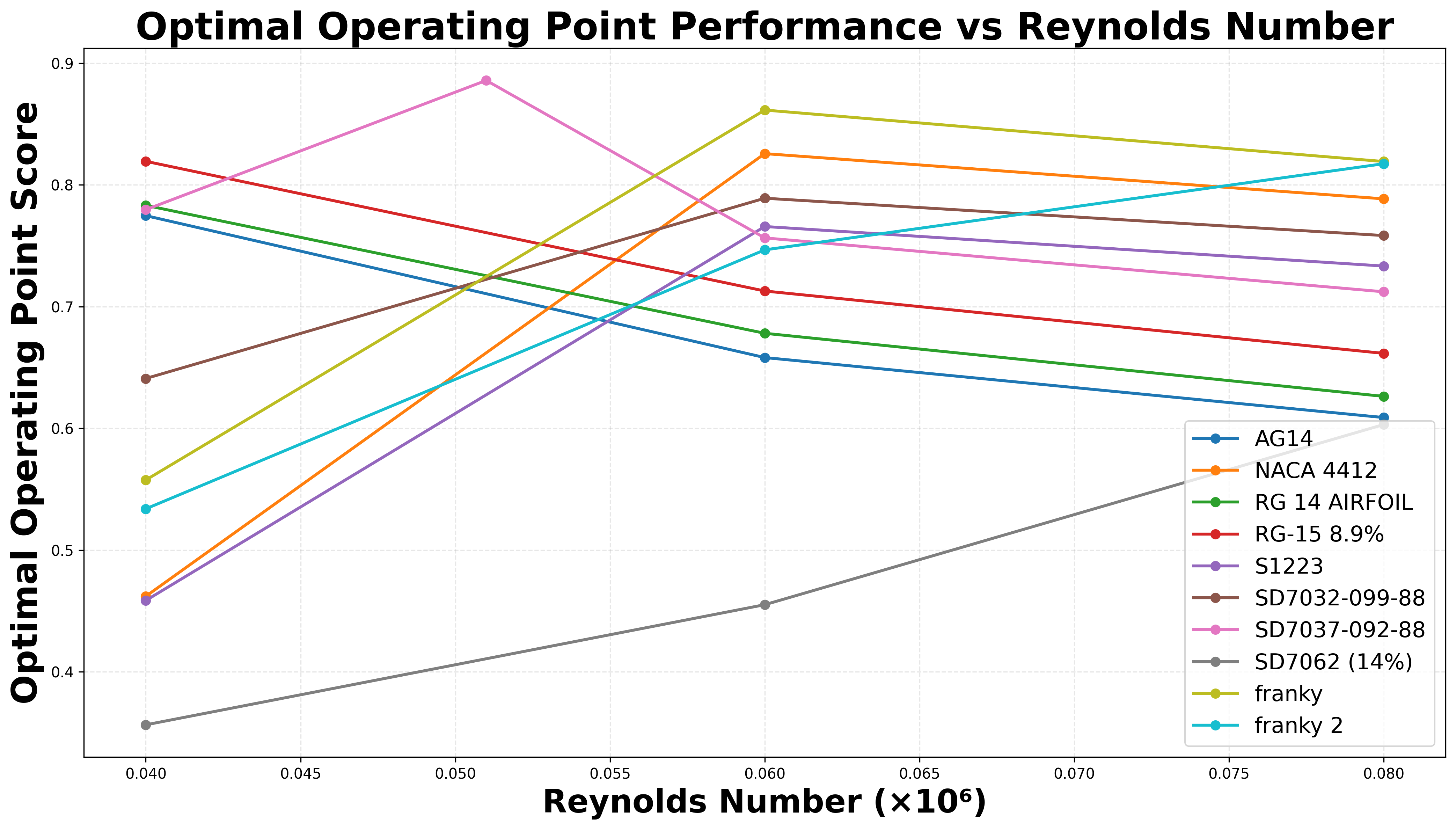}
    \caption{{Optimal operating-point performance score as a function of
Reynolds number for candidate airfoils, using per-Reynolds normalization.}}
    \label{fig:optimal_cost_comp}
\end{figure}

Rather than selecting airfoils based on global peak metrics, a Reynolds-specific optimization strategy was employed. For each Reynolds number, all candidate airfoils and angles of attack were evaluated simultaneously, and performance metrics were normalized within each Reynolds group.
A composite performance score is given in Eq.~\eqref{eq:performance score}.
\begin{equation}
J = 0.50\,\frac{C_l}{C_d} + 0.30\,C_l + 0.20\,C_{p,\min}
\label{eq:performance score}
\end{equation}
where \(C_l/C_d\) represents aerodynamic efficiency, \(C_l\) ensures adequate lift
generation, and \(C_{p,\min}\) penalizes operating points with aggressive pressure
gradients, improving stall margin and robustness.
The inclusion of \(C_{p,\min}\) serves as a heuristic proxy for pressure-gradient severity and stall robustness, rather than a formal stability criterion.
For each airfoil at each Reynolds number, the angle of attack maximizing \(J\) was
selected as the optimal operating point. This procedure yields Reynolds-specific
operating conditions that balance efficiency, thrust generation, and flow robustness, directly informing blade pitch and twist selection. Comparison based on $J$ is shown in Fig ~\ref{fig:optimal_cost_comp}, showing SD7037 outperforming up to 6.0 $\times 10^4$, after which \emph{Franky} performs the best overall.

\subsection{Wing Planform Design for Spinning Hover}
\label{subsec:wing_planform}
\begin{figure}
    \centering
    \includegraphics[width=1\linewidth]{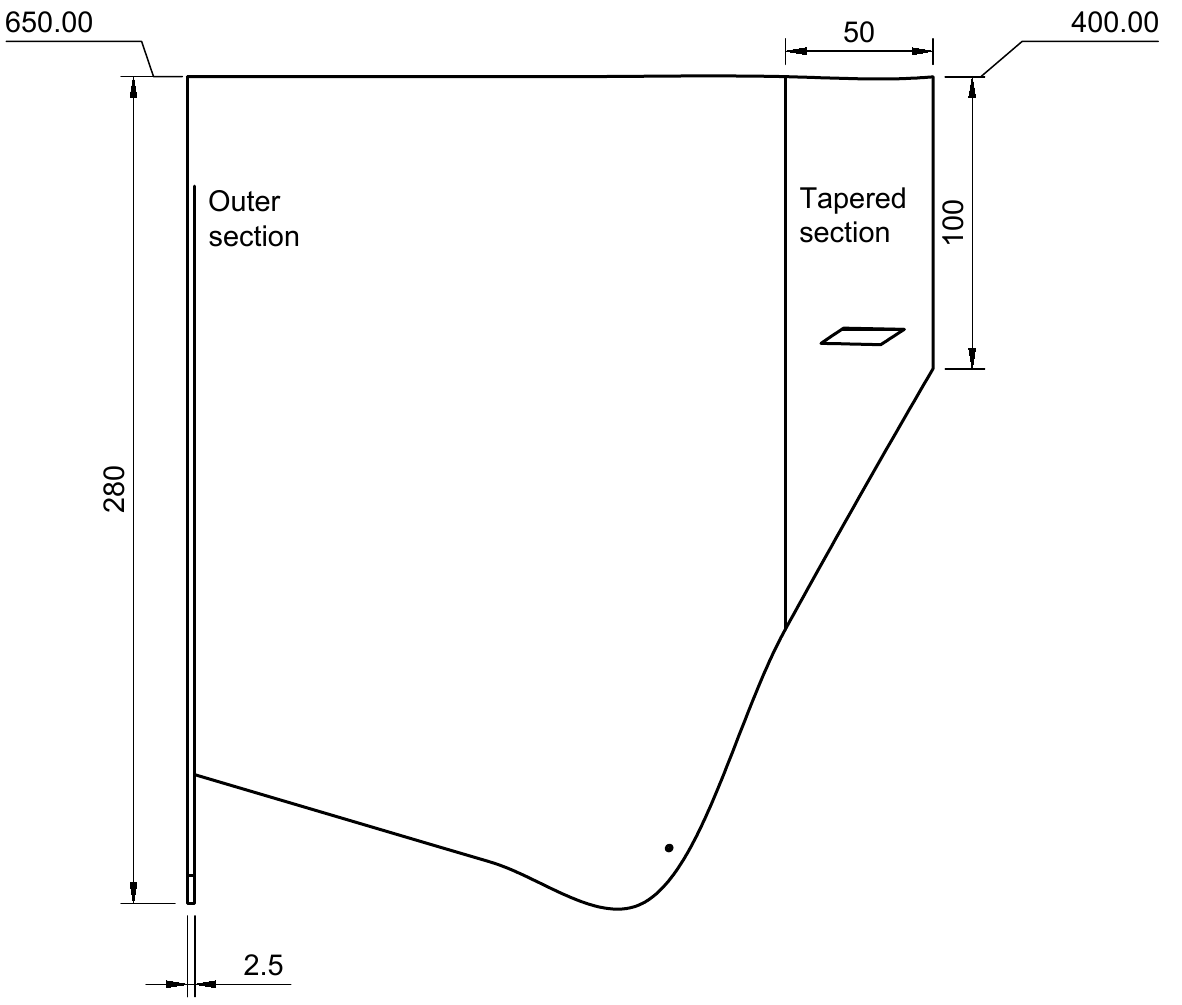}
    \caption{Wing planform (all dimensions in mm)}
    \label{fig:wing_planform}
\end{figure}

As seen in Fig.~\ref{fig:wing_planform}, a short inner transition region was introduced between
$r = 0.40$--$0.45~\mathrm{m}$ to provide a smooth aerodynamic handover
between the hub and the primary lifting sections. This region  is not intended to contribute
significantly to thrust generation. Its purpose is to mitigate root vorticity, flow separation, and three-dimensional interaction effects
at the inner edge of the lifting surface during spinning-hover.

\begin{figure}[t]
    \centering
    \includegraphics[width=0.95\linewidth]{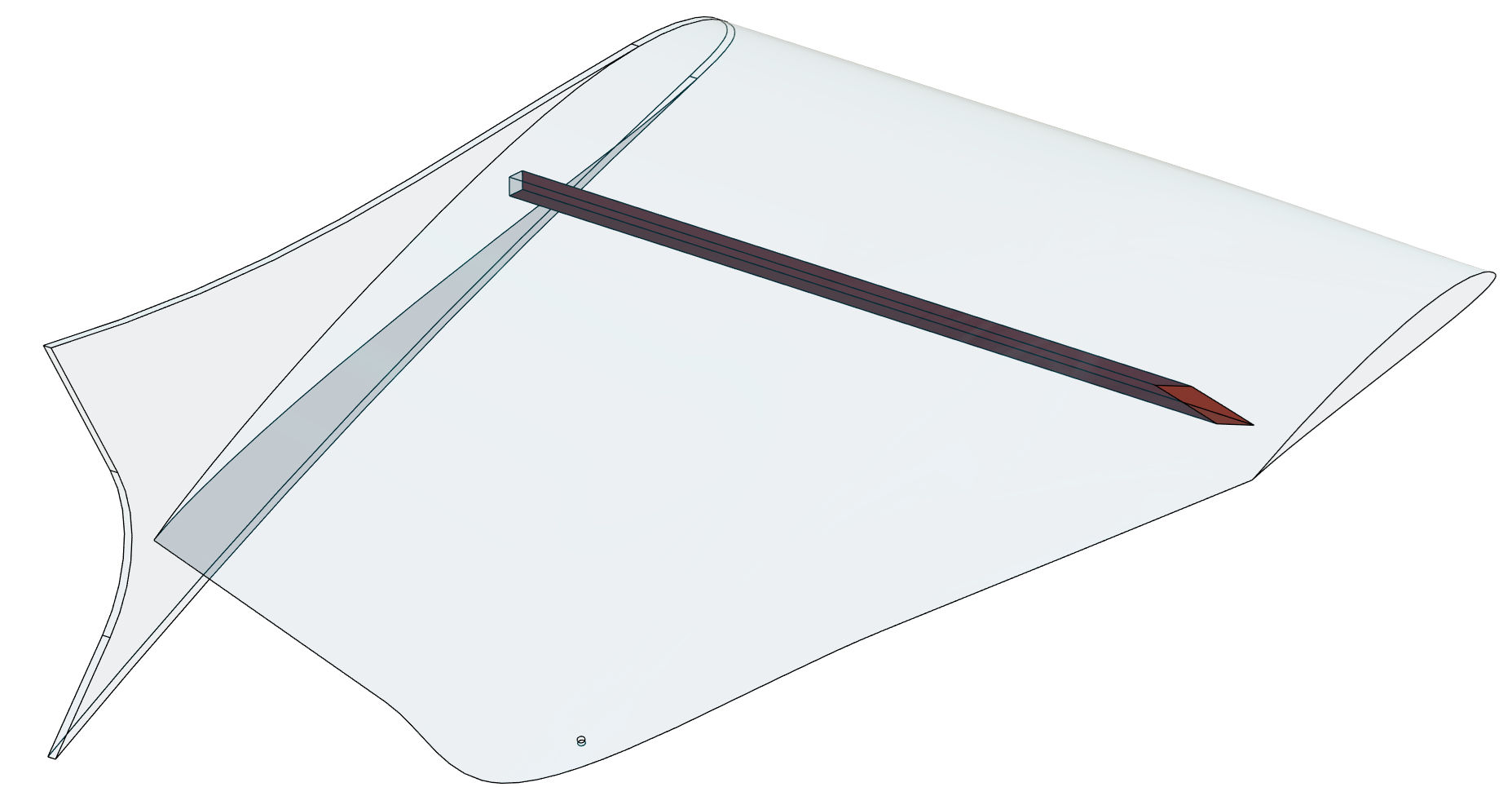}
    \caption{Wing designed for spinning-hover operation.}
    \label{fig:3d_wing}
\end{figure}

The wing planform (Fig \ref{fig:3d_wing}) was generated to match the radial variation in tangential airspeed $V(r)=\omega r$. To reduce profile and induced losses, the chord was tapered toward the inner radius, and inner sections that experience minimal incoming airspeed were omitted. The resulting design, shown in Table~\ref{tab:wing_radial_sections}, targets efficient lift generation in the expected spin-rate band $\omega \approx 8$--$12~\mathrm{rad/s}$.

\begin{table}[t]
    \caption{Radial operating points and section performance used to guide the wing design.}
    \label{tab:wing_radial_sections}
    \renewcommand{\arraystretch}{1.4}
    \centering
    \resizebox{\columnwidth}{!}{%
    \begin{tabular}{c c c c c c c c}
        \hline
        $r$ (m) & $V$ (m/s) & Re ($\times10^4$) & Foil & $\alpha$ ($^\circ$) & $c$ (m) & $C_l$ & $C_l/C_d$ \\
        \hline
        0.40 & 3.2 & 2.2 & SD7037 & 8.6 & 0.10 & 0.83 & 9.3 \\
        0.45 & 3.6 & 4.6 & Franky & 10.5 & 0.19 & 1.00 & 6.3 \\
        0.50 & 4.4 & 8.5 & Franky & 10.0 & 0.29 & 1.79 & 39.6 \\
        0.55 & 4.8 & 8.8 & Franky & 10.0 & 0.27 & 1.79 & 39.6 \\
        0.60 & 5.2 & 9.0 & Franky & 10.0 & 0.26 & 1.79 & 39.6 \\
        0.65 & 5.6 & 9.1 & Franky & 10.0 & 0.24 & 1.79 & 39.6 \\
        \hline
    \end{tabular}}
\end{table}

\section{Results and Discussions}
\subsection{MuJoCo  Simulation}
\label{subsec:simulation_results}
To evaluate the feasibility of spinning hover and quantify potential efficiency benefits,
a coupled simulation framework was developed that integrates
(i) a rigid-body multirotor vehicle model with tilted motors and
(ii) blade-element momentum theory (BEMT) for the rotating lifting surfaces.
Wing aerodynamics are computed using Reynolds-number-dependent airfoil polars,
while a PID altitude controller regulates total motor thrust to maintain hover.

\subsubsection*{Modeling assumptions and scope}
The simulation adopts several idealizations that define the scope of the results.
Wing aerodynamics are modeled as quasi-steady, with two-dimensional airfoil polars
derived from XFLR5 applied to rotating three-dimensional wing sections.
Dynamic stall, azimuthal force variation, and wing--wing aerodynamic interaction
are neglected. Induced flow is assumed axis-symmetric with a simplified tip-loss
correction; wake contraction, ground effect, and wake re-ingestion are not modeled.

Motors and propellers are represented as ideal actuator disks with constant efficiency
$\eta_m = 0.65$, independent of thrust level, rotational speed, or inflow conditions.
Propeller--wing aerodynamic interaction is neglected. Mechanical losses
(e.g., bearing friction, structural damping, vibration) and electrical losses
(e.g., ESCs and wiring) are omitted.

Because identical modeling assumptions are applied to both the spinning-wing
configuration and the motor-only baseline, the simulation enables meaningful
\emph{relative} comparisons of electrical power demand. Absolute power values
should therefore be interpreted as idealized estimates. While unmodeled losses
are expected to reduce real-world efficiency gains relative to the predictions
reported here, the underlying trend—that aerodynamic lift augmentation reduces
propeller thrust requirements - remains consistent.

\subsubsection*{Torque balance and spin equilibrium}
\begin{figure}[t]
    \centering
    \includegraphics[width=0.95\linewidth]{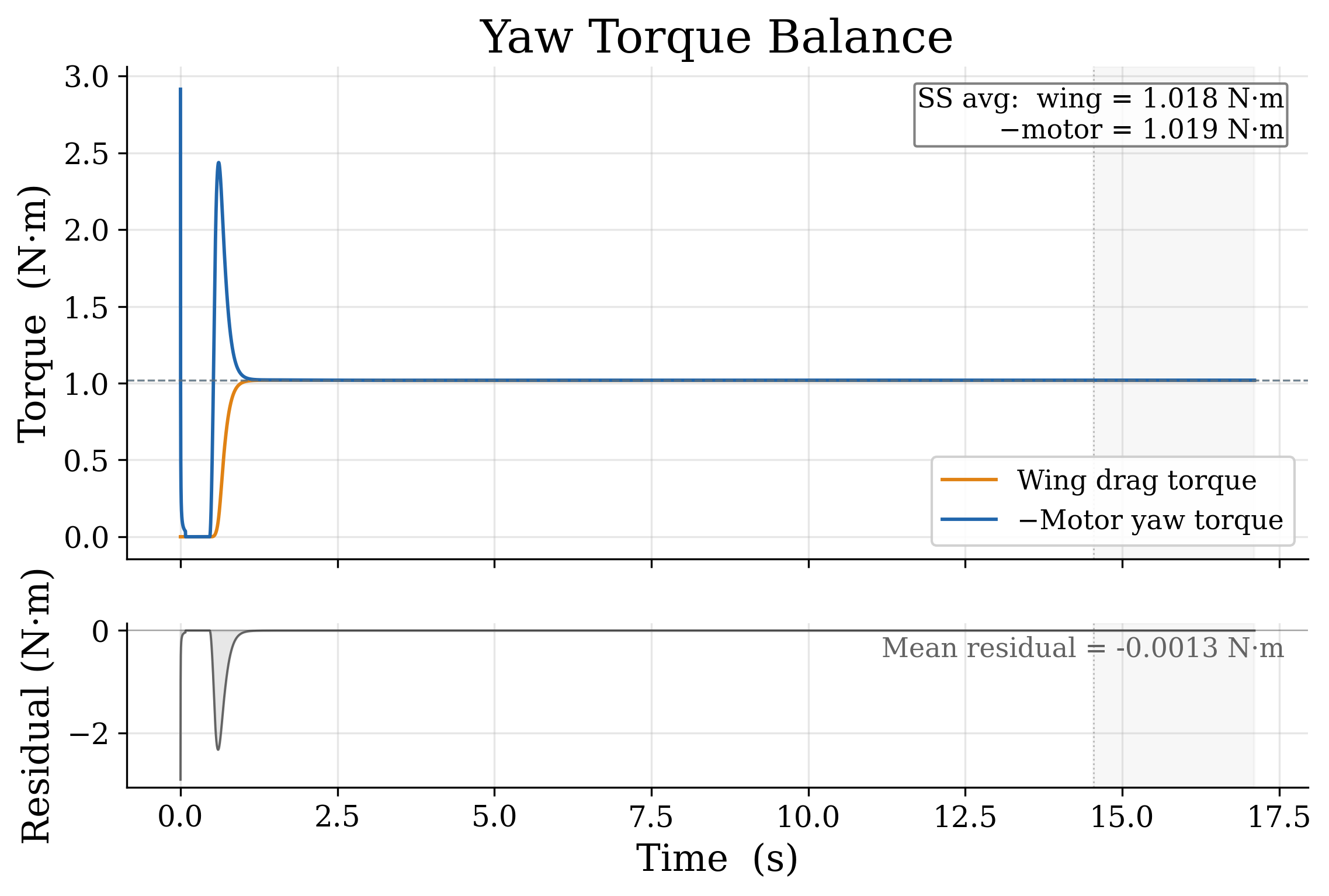}
    \caption{Motor yaw torque and wing drag torque with component breakdown.
    The residual torque remains below 1\% at steady state.}
    \label{fig:sim_torque_balance}
\end{figure}

Fig~\ref{fig:sim_torque_balance} shows convergence to a steady spinning
equilibrium within approximately 2~s. At steady state, the motor-generated yaw
torque balances the aerodynamic drag torque of the rotating wings,
\begin{equation}
\tau_{\text{motor}} \approx \tau_{\text{wing}},
\end{equation}
with steady-state values of 1.033~N$\cdot$m and 1.043~N$\cdot$m, respectively,
indicating agreement to within 1\% and a stable constant-spin operating condition.
The motor yaw torque is dominated by thrust-induced torque from the tilted motors set to $44^\circ$ 
($\sim$98.5\%), with propeller reaction torque contributing only $\sim$1.5\%. The $44^\circ$ tilt is larger than the hardware configuration due to the simplified motor model underestimating torque; a higher cant angle is therefore required in simulation to achieve comparable steady spin rates.

\subsubsection*{Lift distribution}
\begin{figure}[t]
    \centering
    \includegraphics[width=0.95\linewidth]{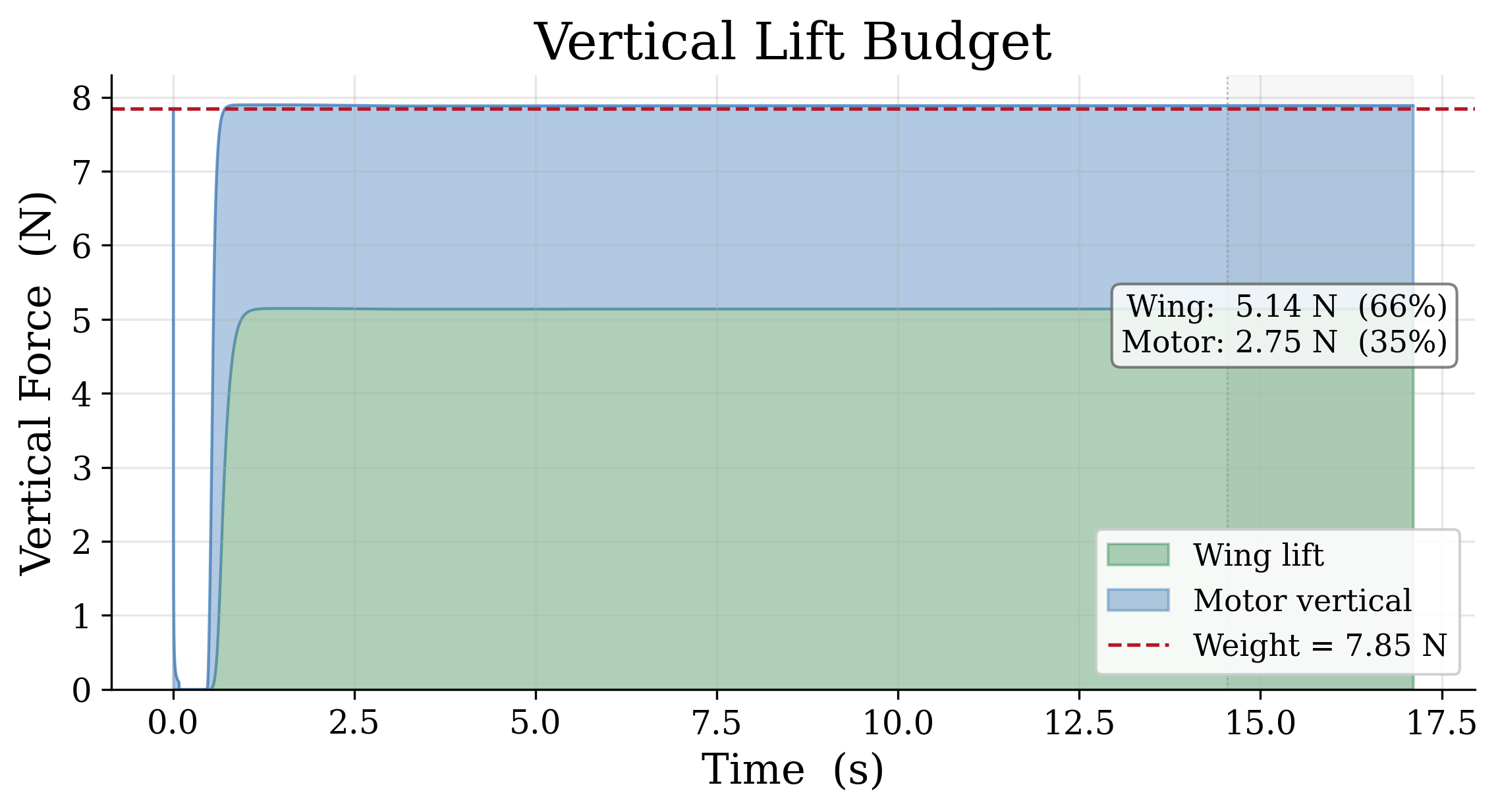}
    \caption{Stacked area plot of vertical force contributions during steady hover.
    The rotating wings supply 67.2\% of the lift, with motors providing the remainder.}
    \label{fig:sim_lift_budget}
\end{figure}

The vertical force balance during hover is given by
\begin{equation}
T_{\text{motor,vert}} + L_{\text{wing}} = W.
\end{equation}
Figure~\ref{fig:sim_lift_budget} shows the steady-state vertical force
distribution during hover. At steady state, the rotating wings generate 5.27~N (67.2\%) of the total vehicle weight of 7.85~N, while the motors provide the remaining 2.58~N (32.8\%). This approximately 2:1 wing-to-motor lift ratio significantly reduces the thrust required from the propellers and constitutes the primary mechanism driving the predicted power reduction.

\subsubsection*{Wing aerodynamic performance}
The system converges to an effective wing lift-to-drag ratio of approximately 2.5 across the span.
The resulting steady-state aerodynamic drag torque of 1.043~N$\cdot$m sets the yaw torque that must be supplied by the motors to sustain rotation.

\subsubsection*{Altitude and yaw rate convergence}
\begin{figure}[t]
    \centering
    \includegraphics[width=0.95\linewidth]{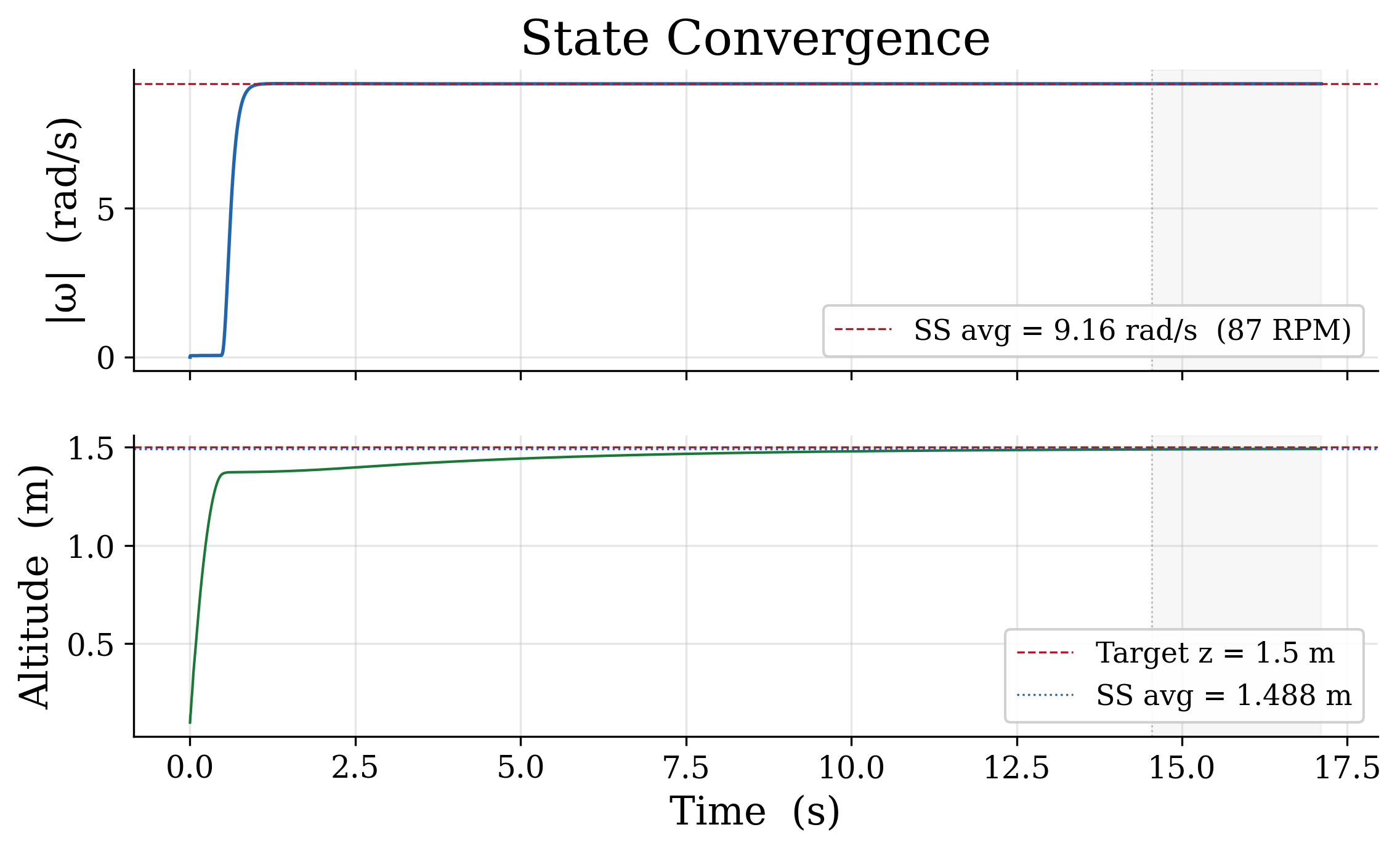}
    \caption{Time histories of yaw rate and altitude showing convergence
    to steady spinning hover.}
    \label{fig:sim_state_convergence}
\end{figure}

Figure~\ref{fig:sim_state_convergence} shows the convergence of altitude to  1.473~m (target altitude: 1.5~m, steady-state error: 0.008~m) and yaw rate
stabilizes to 9.27~rad/s (88.5~RPM).

\subsubsection*{Power analysis}
The power budget provides an idealized estimate of efficiency gains by accounting
for the dominant energy flows in steady hover. The aerodynamic power dissipated
by the rotating wings is given below
\begin{equation}
P_{\text{wing}} = \tau_{\text{wing}} \, \omega,
\label{eq:power dissipation}
\end{equation}
which evaluates to approximately 9.7~W. Using momentum theory, the induced velocity associated with the vertical motor
thrust is
\begin{equation}
v_{i,\text{motor}} =
\sqrt{\frac{T_{\text{motor,vert}}}{2 \rho A_{\text{motor}}}},
\end{equation}
where $T_{\text{motor,vert}}$ (N) is the total vertical motor thrust and 
$A_{\text{motor}}$ (m$^2$) is the total propeller disk area, yielding an induced velocity of 6.82~m/s. The corresponding electrical power required to generate vertical thrust is
\begin{equation}
P_{\text{motor,thrust}} =
\frac{T_{\text{motor,vert}} \, v_{i,\text{motor}}}{\eta_m},
\end{equation}
Which evaluates to 37.6~W. To sustain steady rotation, the motors must additionally supply yaw power given by
\begin{equation}
P_{\text{motor,yaw}} =
\frac{\tau_{\text{motor}} \, \omega}{\eta_m},
\end{equation}
resulting in an electrical yaw power of approximately 14.7~W.

\begin{figure}[t]
    \centering
    \includegraphics[width=0.95\linewidth]{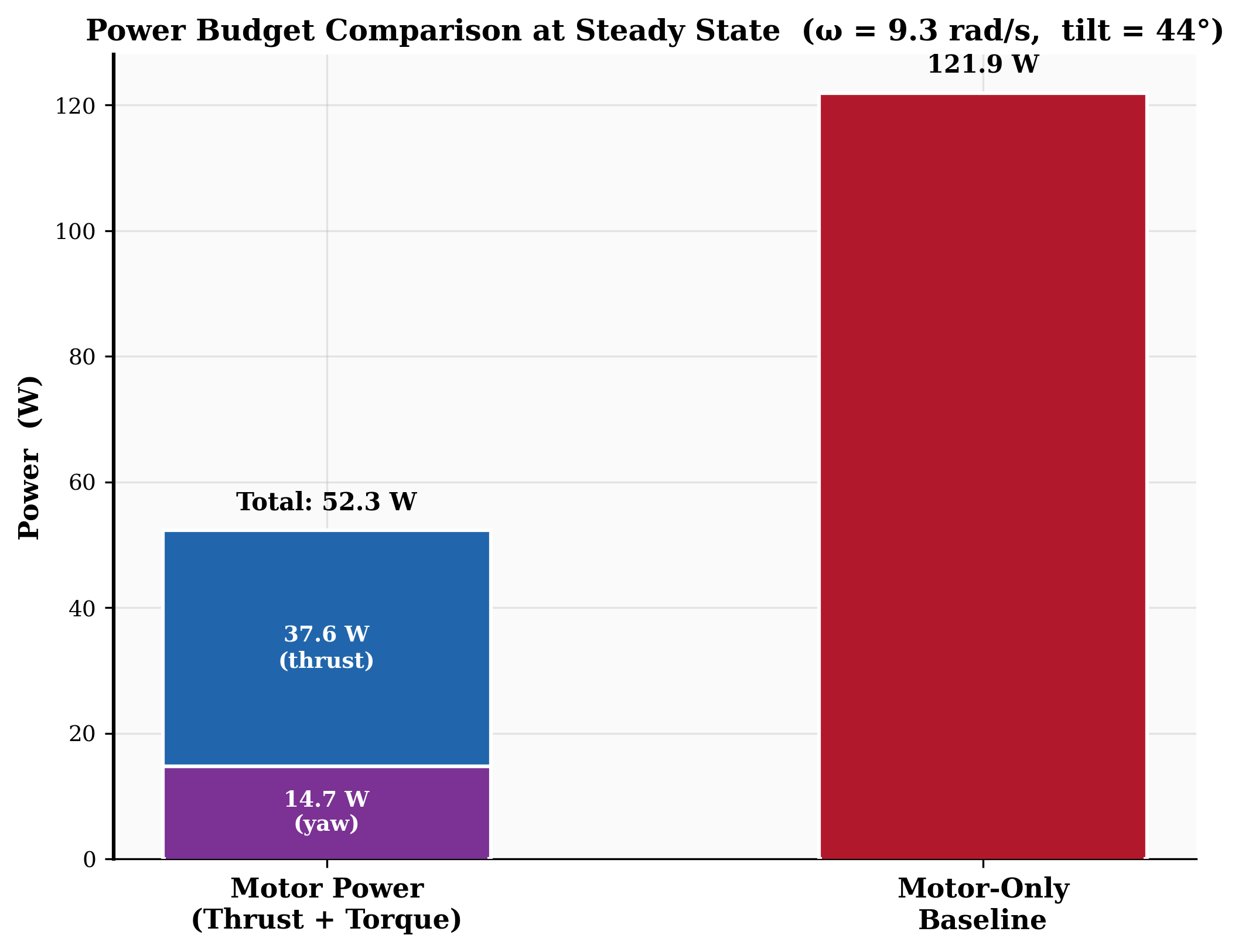}
    \caption{Electrical power breakdown illustrating a 57\% idealized reduction
    relative to a motor-only baseline.}
    \label{fig:sim_power_budget}
\end{figure}

The total steady-state electrical power draw is therefore 52.3~W. Note that the
aerodynamic wing power $P_{\text{wing}}$ represents energy transferred to the
airflow and is already accounted for within $P_{\text{motor,yaw}}$.
For comparison, a conventional motor-only quadcopter operating in hover requires
\begin{equation}
P_{\text{baseline}} =
\frac{W \, v_{i,\text{baseline}}}{\eta_m},
\qquad
v_{i,\text{baseline}} =
\sqrt{\frac{W}{2 \rho A_{\text{motor}}}},
\end{equation}
which evaluates to a baseline electrical power of 122.0~W. Under the idealized modeling assumptions described above, the resulting electrical power estimate corresponds to a 57.1\% reduction relative to a motor-only hover baseline. This value represents an upper-bound prediction under quasi-steady aerodynamics and constant motor efficiency assumptions. Figure~\ref{fig:sim_power_budget} summarizes the electrical power breakdown
for the spinning configuration and the motor-only baseline.

\begin{figure}
    \centering
    \includegraphics[width=\linewidth]{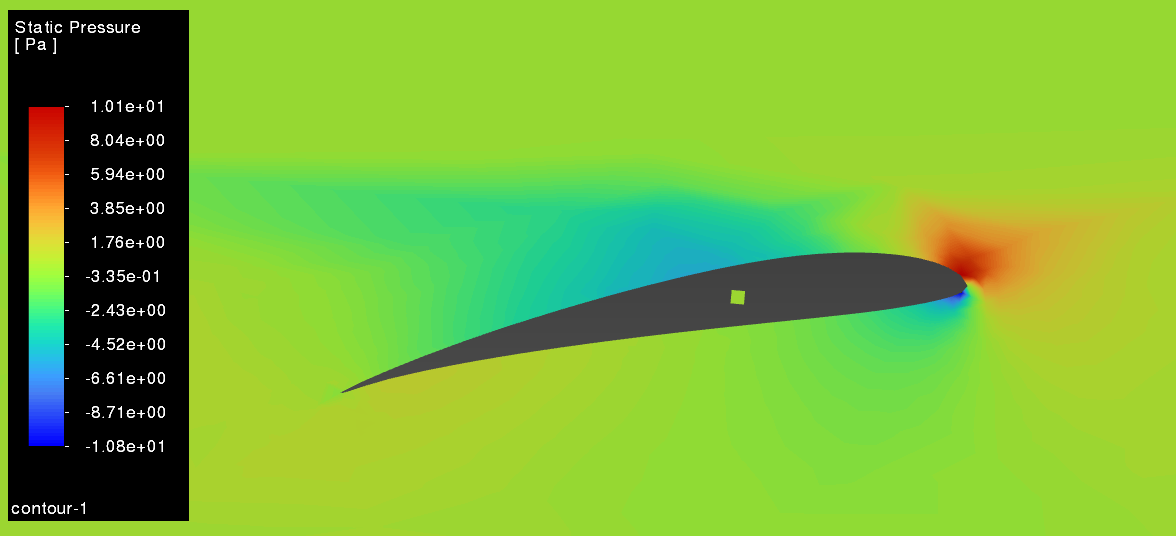}
    \caption{ANSYS Fluent static pressure contour around a wing cross-section, illustrating the local pressure field under the aerodynamic effect of the other wings}
    \label{fig:cfd_contour}
\end{figure}
Overall, the simulation demonstrates that intentional spinning hover can
substantially reduce propeller power requirements by transferring the majority
of lift generation to aerodynamic wings. The reported power reduction represents
an upper bound under idealized assumptions. Real-world performance is expected
to be reduced by aerodynamic interaction effects, non-uniform inflow, motor
efficiency variation, and mechanical losses, underscoring the need for
experimental validation.

\subsection{CFD Cross-Validation}

\noindent A steady-state CFD analysis was performed in ANSYS Fluent using a multiple reference frame (MRF) formulation at a rotational speed of $\omega = 10~\mathrm{rad/s}$. The low-Reynolds-number $k$--$\omega$ SST turbulence model was employed to capture boundary-layer behavior in the $Re \sim 10^5$ regime. The simulation considered the complete four-wing assembly in rotation, while excluding propeller-induced flow to isolate lift generated solely by the rotating aerodynamic surfaces. The integrated lift was $L_{\text{CFD}} = 4.50~\mathrm{N}$. The static pressure contour shown in Fig.~\ref{fig:cfd_contour} represents a cross-sectional slice of a representative wing, illustrating the pressure differential responsible for lift generation. Unlike the CFD model, MuJoCo   simulations do not capture three-dimensional aerodynamic blade–blade interactions or detailed flow effects; instead, they approximate lift contributions at a system level. Consequently, CFD provides a more accurate bound than MuJoCo.
\subsection{Hardware Hover Performance}
\label{subsec:hardware_hover}
\begin{figure}[t]
    \centering
    \includegraphics[width=0.75\linewidth]{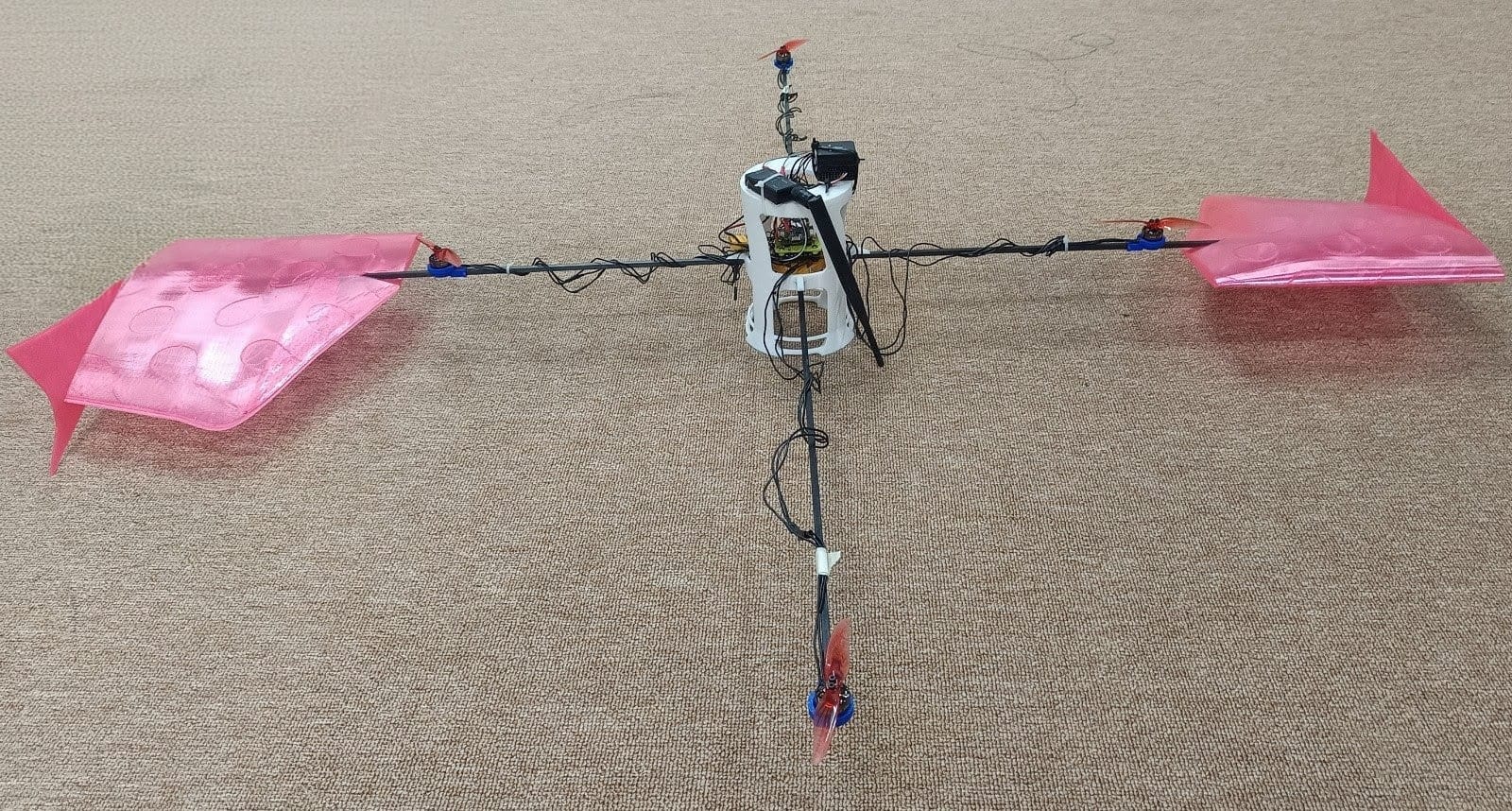}
    \caption{Configuration I: Orthogonal wing–motor layout with asymmetric motor tilt.}
    \label{fig:config1}
\end{figure}

\begin{figure}[t]
    \centering
    \includegraphics[width=0.75\linewidth]{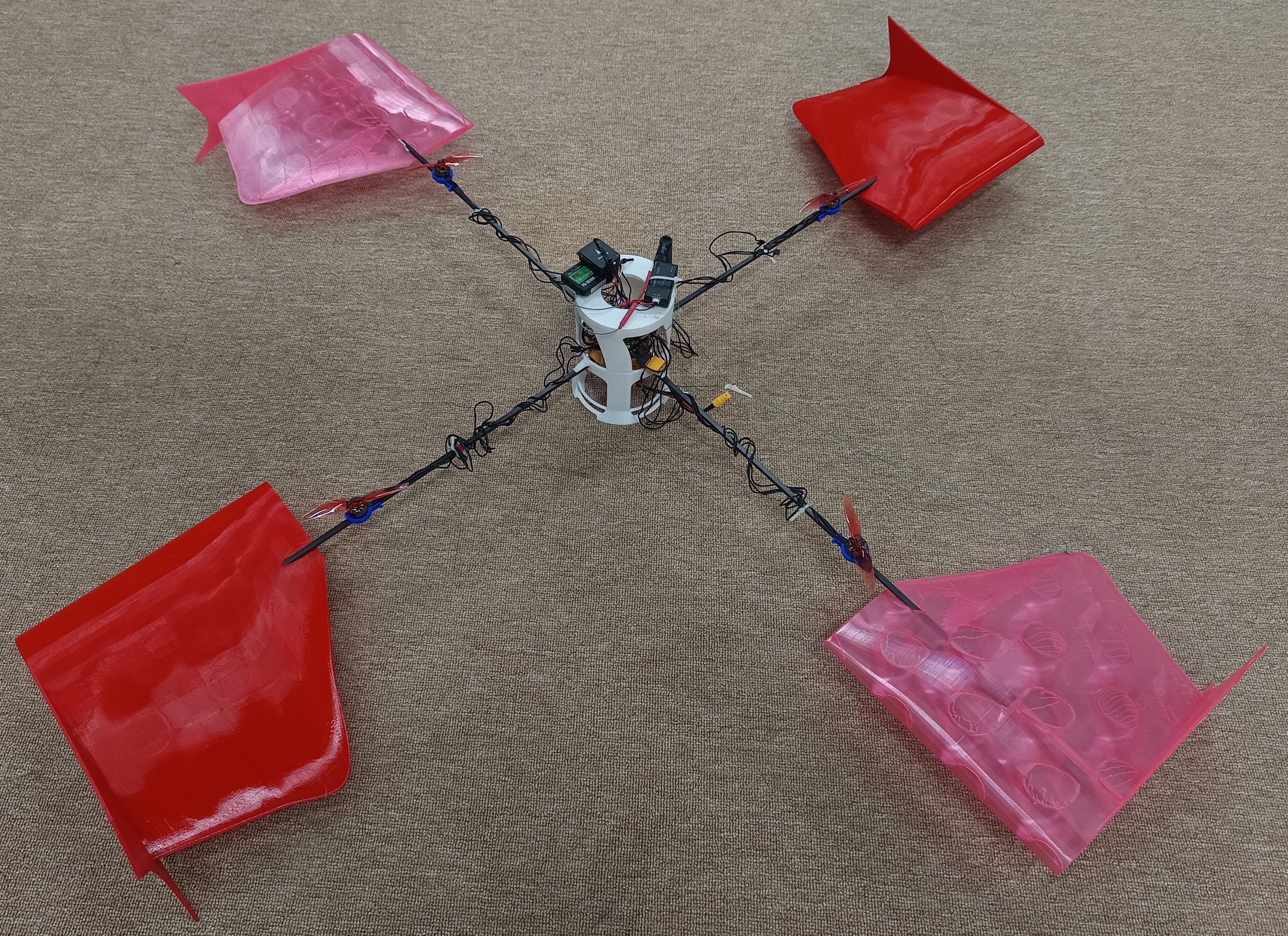}
    \caption{Configuration II: Fully tilted motor layout with distributed wings. Representative flight tests for all configurations are shown in the supplementary video available at:
\url{https://youtu.be/q07RIcLkxVw}.}
    \label{fig:config2}
\end{figure}

\noindent  Hover flight tests were conducted across four hardware configurations to quantify the aerodynamic contribution of rotating wings during sustained spinning hover. The evaluated cases comprise spinning and non-spinning variants of both the asymmetric two-wing configuration (Configuration~I) and the symmetric four-wing configuration (Configuration~II), each tested at identical vehicle mass.
The tested hardware platforms are shown in Fig.~\ref{fig:config1} and Fig.~\ref{fig:config2}. Configuration~I employs an orthogonal two-wing layout with asymmetric motor tilt, whereas Configuration~II uses a symmetric four-wing arrangement with fully tilted motors. For each flight, a stabilized hover interval was identified by excluding transient ascent and descent phases and selecting steady-state motor command regions. Wing lift was inferred from vertical force equilibrium,
\begin{equation}
L_{\text{wing}} = m g - T_{\text{motor}}\cos\theta,
\end{equation}
where \(m\) is the vehicle mass, \(g\) is gravitational acceleration, \(T_{\text{motor}}\) is the total motor thrust required for hover, and \(\theta\) is the motor tilt angle. Table~\ref{tab:hover_performance} summarizes the measured hover performance across all configurations, and Table~\ref{tab:lift_comparison} summarizes across the various result sections.
\begin{table*}[t]
\centering
\caption{Hover performance metrics across spinning and stationary configurations.}
\label{tab:hover_performance}
\renewcommand{\arraystretch}{1.4}
\begin{tabular}{l c c c c c c}
\hline
Configuration & Mass (g) & DSHOT & Motor Thrust (g) & Wing Lift (g) & \% Weight Lifted & Per-Wing Lift (g) \\
\hline
Stationary 4-Wing & 960 & 1809 & 960 & 0 & 0.0 & --- \\
Spinning 4-Wing (+10°) & 960 & 1487 & 758 & 214 & 22.3 & 53.5 \\
Stationary 2-Wing & 658 & 1339 & 658 & 0 & 0.0 & --- \\
Spinning 2-Wing & 658 & 1133 & 511 & 147 & 22.3 & 73.4 \\
\hline
\end{tabular}
\end{table*}

\begin{figure}[t]
    \centering
    \includegraphics[width=\linewidth]{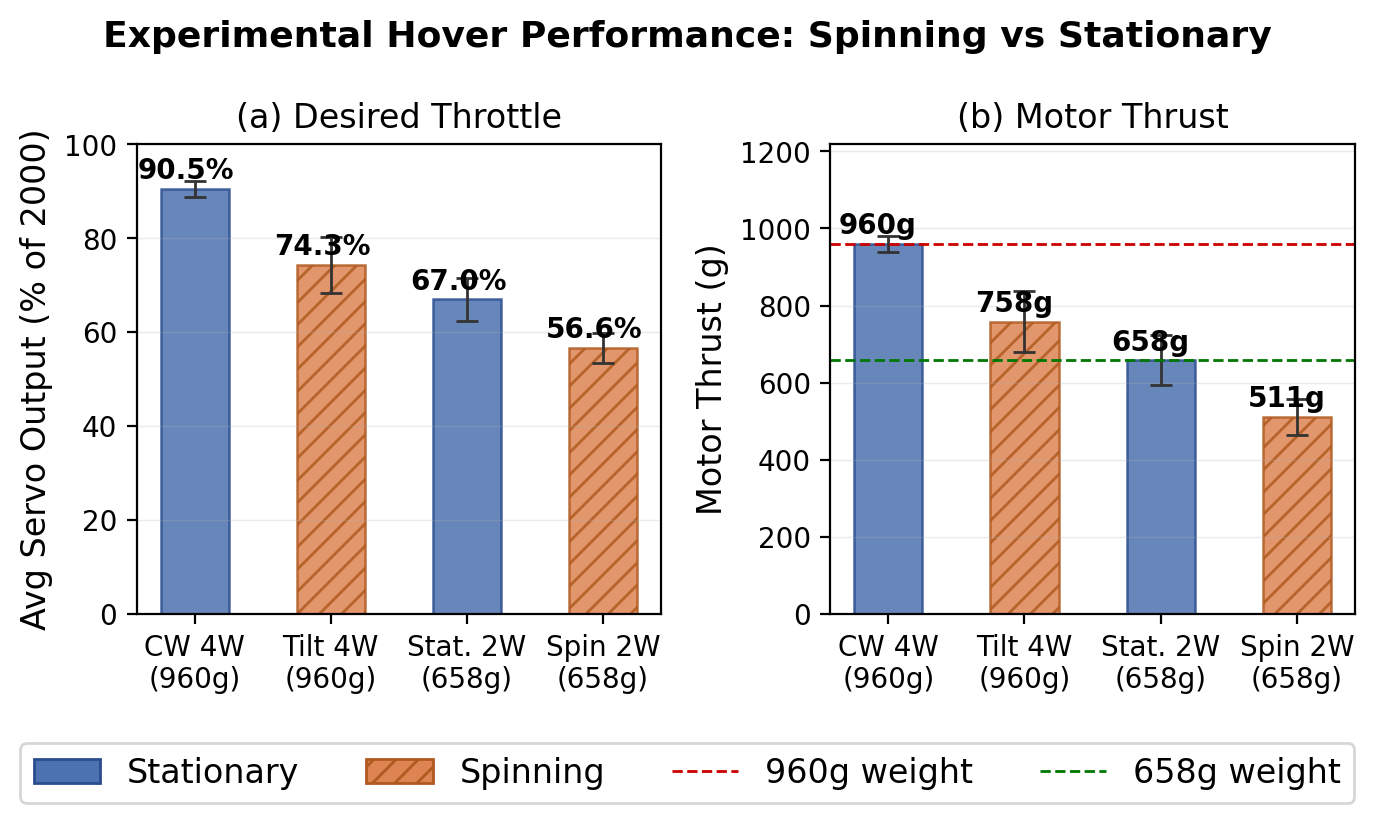}
    \caption{Comparison of thrust production and hover efficiency for spinning and stationary configurations under identical payload conditions. Each reported value represents the mean over $N=5$ independent hover trials, each consisting of a 10~s steady-state window. Error bars in the figures denote one standard deviation across trials.}
    \label{fig:thrust_efficiency_comparison}
\end{figure}

Figure~\ref{fig:thrust_efficiency_comparison} compares thrust production at hover across spinning and stationary configurations. Across both airframe geometries, spinning configurations consistently required lower motor thrust to maintain hover than their stationary counterparts. In the four-wing configuration, steady hover was achieved at a mean command of 1487 DSHOT units—an 18\% reduction relative to the stationary four-wing baseline. At this operating point, the motors generated 758~g of total thrust, corresponding to 746~g of vertical force after accounting for the 10° tilt. For a 960~g vehicle, the remaining 214~g (22.3\%) of lift was supplied by the rotating wings which is almost as much as an additional motor. Consequently, the motors provided approximately 21\% less thrust than would be required in a motor-only hover condition.

A comparable fractional lift contribution was observed in the two-wing configuration. The spinning two-wing platform required a 15\% lower motor command than its stationary baseline, with the rotating wings supplying 147~g of lift, corresponding to 22.3\% of vehicle weight. Notably, the per-wing lift in the two-wing case (73.4~g per wing) exceeded that of the four-wing configuration (53.5~g per wing), despite similar total lift fractions.

The reduced per-wing lift observed in the four-wing spinning configuration is attributed to increased aerodynamic interaction and wake interference arising from the higher blade count. With four lifting surfaces operating at similar radii, wake overlap and induced velocity effects reduce the effective angle of attack experienced by downstream sections, thereby lowering the lift generated per wing. 

\begin{figure}
    \centering
    \includegraphics[width=1\linewidth]{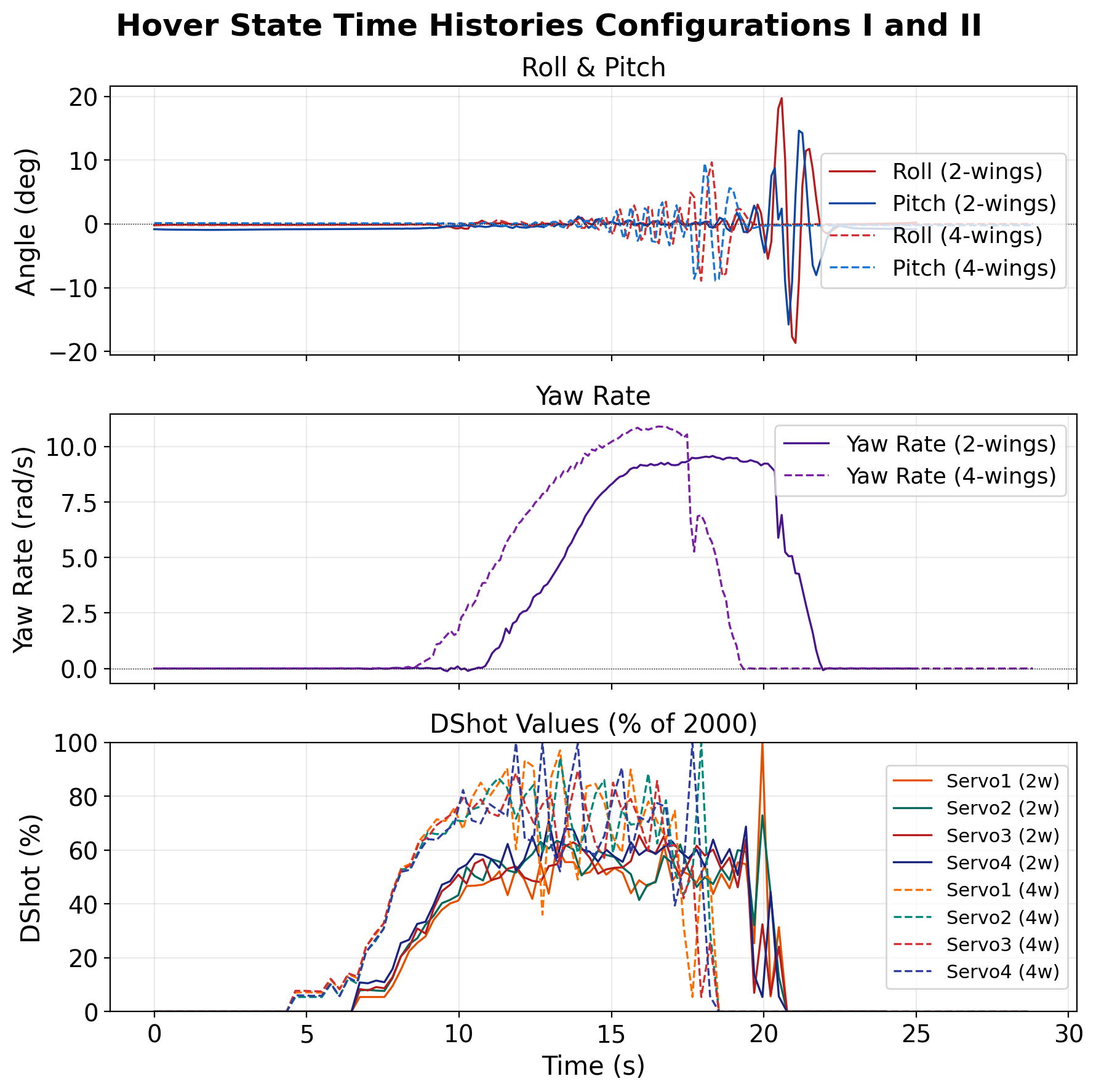}
    \caption{Experimental Roll, pitch stability, yaw rate and DSHOT history of configuration 1 and 2}
    \label{fig:hist}
\end{figure}

The stability and yaw-rate and throttle over flight of configuration 1 and 2 can be seen in Figure~\ref{fig:hist}.

\begin{table}[t]
\centering
\caption{Comparison of wing lift predictions across modeling and experiment.}
\renewcommand{\arraystretch}{1.3}
\begin{tabular}{l c}
\hline
Method & Total Lift (N) \\
\hline
BEMT Simulation (9.27 rad/s) & 5.27 \\
CFD (MRF, $k$--$\omega$ SST) & 4.50 \\
Hardware (experimental results) & 2.10 \\
per motor thrust (configuration 2) & 2.35\\
\hline
\end{tabular}
\label{tab:lift_comparison}
\end{table}
\section{Conclusion and Future Work}

The results demonstrate that intentional spinning hover can reduce propeller thrust demand by transferring a portion of lift generation to rotating aerodynamic surfaces, with simulations establishing an upper bound and hardware experiments confirming practical feasibility. Notably, in spinning configurations the wings generate lift comparable to that produced by one of the four motors during stationary hover, highlighting their meaningful aerodynamic contribution. Future work will primarily focus on fault-tolerant control, particularly exploiting rotational dynamics to sustain stability and lift under partial motor failure. Emphasis will be placed on control strategies that leverage passive aerodynamic stabilization while ensuring robustness under actuator degradation, positioning spinning hover as a viable fault-tolerant operating regime for multirotor platforms.

\section{Acknowledgments}
The authors acknowledge the support provided by MeitY, Govt. of India, under the project "Capacity Building for Human Resource Development in Unmanned Aircraft System (Drone and Related Technology)."

\bibliographystyle{IEEEtran}
\bibliography{references}

@inproceedings{mueller_stability_2014,
	address = {Hong Kong, China},
	title = {Stability and control of a quadrocopter despite the complete loss of one, two, or three propellers},
	isbn = {978-1-4799-3685-4},
	url = {http://ieeexplore.ieee.org/document/6906588/},
	doi = {10.1109/ICRA.2014.6906588},
	language = {en},
	urldate = {2025-09-05},
	booktitle = {2014 {IEEE} {International} {Conference} on {Robotics} and {Automation} ({ICRA})},
	publisher = {IEEE},
	author = {Mueller, Mark W. and D'Andrea, Raffaello},
	month = may,
	year = {2014},
	pages = {45--52},
}

@inproceedings{cai_cooperative_2022,
	address = {Philadelphia, PA, USA},
	title = {Cooperative {Modular} {Single} {Actuator} {Monocopters} {Capable} of {Controlled} {Passive} {Separation}},
	copyright = {https://doi.org/10.15223/policy-029},
	isbn = {978-1-7281-9681-7},
	url = {https://ieeexplore.ieee.org/document/9812182/},
	doi = {10.1109/ICRA46639.2022.9812182},
	language = {en},
	urldate = {2025-09-05},
	booktitle = {2022 {International} {Conference} on {Robotics} and {Automation} ({ICRA})},
	publisher = {IEEE},
	author = {Cai, Xinyu and Win, Shane Kyi Hla and Win, Luke Soe Thura and Sufiyan, Danial and Foong, Shaohui},
	month = may,
	year = {2022},
	pages = {1989--1995},
}

@article{freddi_feedback_2011,
	title = {A {Feedback} {Linearization} {Approach} to {Fault} {Tolerance} in {Quadrotor} {Vehicles}},
	volume = {44},
	copyright = {https://www.elsevier.com/tdm/userlicense/1.0/},
	issn = {14746670},
	url = {https://linkinghub.elsevier.com/retrieve/pii/S1474667016444678},
	doi = {10.3182/20110828-6-IT-1002.02016},
	language = {en},
	number = {1},
	urldate = {2025-09-05},
	journal = {IFAC Proceedings Volumes},
	author = {Freddi, Alessandro and Lanzon, Alexander and Longhi, Sauro},
	month = jan,
	year = {2011},
	pages = {5413--5418},
}

@article{lanzon_flight_2014,
	title = {Flight {Control} of a {Quadrotor} {Vehicle} {Subsequent} to a {Rotor} {Failure}},
	volume = {37},
	issn = {0731-5090, 1533-3884},
	url = {https://arc.aiaa.org/doi/10.2514/1.59869},
	doi = {10.2514/1.59869},
	language = {en},
	number = {2},
	urldate = {2025-09-05},
	journal = {Journal of Guidance, Control, and Dynamics},
	author = {Lanzon, Alexander and Freddi, Alessandro and Longhi, Sauro},
	month = mar,
	year = {2014},
	pages = {580--591},
}

@article{orsag_spincopter_2013,
	title = {Spincopter {Wing} {Design} and {Flight} {Control}},
	volume = {70},
	copyright = {http://www.springer.com/tdm},
	issn = {0921-0296, 1573-0409},
	url = {http://link.springer.com/10.1007/s10846-012-9725-2},
	doi = {10.1007/s10846-012-9725-2},
	language = {en},
	number = {1-4},
	urldate = {2025-09-05},
	journal = {Journal of Intelligent \& Robotic Systems},
	author = {Orsag, Matko and Cesic, Josip and Haus, Tomislav and Bogdan, Stjepan},
	month = apr,
	year = {2013},
	pages = {165--179},
}

@inproceedings{low_design_2017,
	address = {Singapore, Singapore},
	title = {Design and dynamic analysis of a {Transformable} {Hovering} {Rotorcraft} ({THOR})},
	isbn = {978-1-5090-4633-1},
	url = {http://ieeexplore.ieee.org/document/7989755/},
	doi = {10.1109/ICRA.2017.7989755},
	language = {en},
	urldate = {2025-09-05},
	booktitle = {2017 {IEEE} {International} {Conference} on {Robotics} and {Automation} ({ICRA})},
	publisher = {IEEE},
	author = {Low, Jun En and Win, Luke Thura Soe and Shaiful, Danial Sufiyan Bin and Tan, Chee How and Soh, Gim Song and Foong, Shaohui},
	month = may,
	year = {2017},
	pages = {6389--6396},
}

@article{sun_autonomous_2021,
	title = {Autonomous {Quadrotor} {Flight} {Despite} {Rotor} {Failure} {With} {Onboard} {Vision} {Sensors}: {Frames} vs. {Events}},
	volume = {6},
	copyright = {https://ieeexplore.ieee.org/Xplorehelp/downloads/license-information/IEEE.html},
	issn = {2377-3766, 2377-3774},
	shorttitle = {Autonomous {Quadrotor} {Flight} {Despite} {Rotor} {Failure} {With} {Onboard} {Vision} {Sensors}},
	url = {https://ieeexplore.ieee.org/document/9312462/},
	doi = {10.1109/LRA.2020.3048875},
	language = {en},
	number = {2},
	urldate = {2025-09-05},
	journal = {IEEE Robotics and Automation Letters},
	author = {Sun, Sihao and Cioffi, Giovanni and De Visser, Coen and Scaramuzza, Davide},
	month = apr,
	year = {2021},
	pages = {580--587},
}

@article{win_design_2021,
	title = {Design and control of the first foldable single-actuator rotary wing micro aerial vehicle},
	volume = {16},
	issn = {1748-3182, 1748-3190},
	url = {https://iopscience.iop.org/article/10.1088/1748-3190/ac253a},
	doi = {10.1088/1748-3190/ac253a},
	language = {en},
	number = {6},
	urldate = {2025-09-06},
	journal = {Bioinspiration \& Biomimetics},
	author = {Win, Shane Kyi Hla and Win, Luke Soe Thura and Sufiyan, Danial and Foong, Shaohui},
	month = nov,
	year = {2021},
	pages = {066019},
}

@article{ke2023uniform,
  title={Uniform Passive Fault-Tolerant Control of a Quadcopter With One, Two, or Three Rotor Failure},
  author={Ke, Chenxu and Cai, Kai-Yuan and Quan, Quan},
  journal={IEEE Transactions on Robotics},
  volume={39},
  number={6},
  pages={4827--4846},
  year={2023},
  publisher={IEEE}
}

@article{bai2022bioinspired,
  title={A bioinspired revolving-wing drone with passive attitude stability and efficient hovering flight},
  author={Bai, Songnan and He, Qilin and Chirarattananon, Pakpong},
  journal={Science Robotics},
  volume={7},
  number={66},
  pages={eabg5913},
  year={2022},
  publisher={American Association for the Advancement of Science}
}

@inproceedings{hedayatpour2017optimal,
  title={Optimal-power Configurations for Hover Solutions in Mono-spinners},
  author={Hedayatpour, Mojtaba and Mehrandezh, Mehran and Janabi-Sharifi, Farrokh},
  booktitle={2017 IEEE/RSJ International Conference on Intelligent Robots and Systems (IROS)},
  year={2017},
  note={arXiv:1709.07969}
}

@inproceedings{invernizzi2018full,
  title={Full pose tracking for a tilt-arm quadrotor UAV},
  author={Invernizzi, Davide and Giurato, Mattia and Gattazzo, Paolo and Lovera, Marco},
  booktitle={2018 IEEE Conference on Control Technology and Applications (CCTA)},
  pages={1--6},
  year={2018},
  organization={IEEE}
}

\end{document}